\documentclass[11pt]{article}
\usepackage{amsmath,amssymb, abstract}
\usepackage{graphicx}
\usepackage{hyperref}
\usepackage{moreverb,url}
\usepackage{booktabs}
\usepackage{float}
\usepackage{adjustbox}
\usepackage{enumitem}
\usepackage{xurl}
\usepackage{authblk}

\title{Financial Fraud Detection Using Explainable AI and Stacking Ensemble Methods}

\author[1]{Fahad Almalki\thanks{\href{mailto:s44580241@students.tu.edu.sa}{s44580241@students.tu.edu.sa}}}
\author[2]{Mehedi Masud\thanks{\href{mailto:mmasud@tu.edu.sa}{mmasud@tu.edu.sa}}}

\affil[1]{Computer Science Department, College of Computers and Information Technology, Taif University, Al-Hawiyah, Taif 21944, Makkah, Saudi Arabia}
\affil[2]{Department of Computer Science, Taif University, Al-Hawiyah, Taif 21944, Makkah, Saudi Arabia}

\date{May 2025}

\begin{document}

\maketitle

\begin{abstract}
Losses from credit card fraud increased significantly worldwide from \$28.4 billion in 2020 to \$33.5 billion in 2022.  However, traditional machine learning models often prioritize predictive accuracy, often at the expense of model transparency and interpretability. The lack of transparency makes it difficult for organizations to comply with regulatory requirements and gain stakeholders’ trust. In this research, we propose a fraud detection framework that combines a stacking ensemble of well-known gradient boosting models: XGBoost, LightGBM, and CatBoost. In addition, explainable artificial intelligence (XAI) techniques are used to enhance the transparency and interpretability of the model’s decisions. We used SHAP (SHapley Additive Explanations) for feature selection to identify the most important features. Further efforts were made to explain the model’s predictions using Local Interpretable Model-Agnostic Explanation (LIME), Partial Dependence Plots (PDP), and  Permutation Feature Importance (PFI). The IEEE-CIS Fraud Detection dataset, which includes more than 590,000 real transaction records, was used to evaluate the proposed model. The model achieved a high performance with an accuracy of 99\% and an AUC-ROC score of 0.99, outperforming several recent related approaches. These results indicate that combining high prediction accuracy with transparent interpretability is possible and could lead to a more ethical and trustworthy solution in financial fraud detection.
\end{abstract}

%\keywords{Artificial Intelligence, Stacking Ensemble, Machine Learning, Model Interpretability, XGBoost, LightGBM, CatBoost, SHAP, LIME}

\section{Introduction}
Artificial Intelligence (AI) plays a role in fraud prevention, but as these systems become more complex, they often lose transparency. This creates the “black-box” problem, where users and financial institutions cannot clearly trace how a decision was made, creating concerns regarding trust and ethical standards in the use of AI. Despite AI’s potential to reduce operational costs and improve efficiency, the adoption of AI in fraud detection remains relatively low, with only 22\% of businesses utilizing it in 2023 \cite{ibm2024}.

The rapid expansion of digital financial services has emphasized the growing need for both security and transparency. As financial institutions increasingly rely on machine learning models to detect fraud, concerns about the explainability of these models have become more evident. Users and regulators alike are looking for systems that offer high prediction accuracy and provide clear, understandable explanations for the model’s decisions \cite{deloitte2024}. Failure to explain AI output may lead to legal and reputational issues for institutions, especially those operating in high-risk industries such as finance. Regulatory standards stress the importance of giving clear and meaningful reasons for automated decisions \cite{arrieta2020}. Financial institutions are encouraged to incorporate XAI techniques to meet these evolving regulatory requirements \cite{jain2024}.

Although deep learning models are known to perform well in identifying sophisticated fraud patterns, their limited interpretability still presents a serious challenge \cite{gunning2019}. To address the well-established conflict between predictive performance and model interpretability, XAI frameworks have emerged as a viable solution \cite{deloitte2024}. Tools like SHAP and LIME help reveal how input features influence a model’s output, offering clearer insight into its decisions and encouraging greater trust in the system \cite{gunning2019}. By combining ensemble stacking techniques with XAI methods, organizations could develop systems that are not only accurate in detecting fraud but also capable of providing helpful explanations for their recommendations \cite{deloitte2024}.

The increasing losses caused by global credit card fraud, from \$28.4 billion in 2020 to \$33.5 billion in 2022, show the urgent need to design more effective detection mechanisms. These losses are expected to grow further, reaching \$43.47 billion by 2028 \cite{webb2025}. Although machine learning techniques efficiently uncover complex patterns within large datasets, they often provide limited insight into their decision-making processes \cite{deloitte2024}. XAI methods attempt to find a reasonable balance between performance and interpretability by supporting models that are both transparent and accurate \cite{gunning2019}. Recent developments showed that a stacked ensemble combining Random Forest, AdaBoost, and GBDT classifiers was able to detect financial fraud more effectively than any individual model alone \cite{veigas2021}.

As the number of digital transactions continues to increase and public concerns regarding data use become more prominent, the need for ethical and explainable data processing systems also grows. The implementation of XAI helps organizations comply with data privacy laws such as GDPR \cite{arrieta2020}, and it plays a significant role in maintaining consumer trust. According to recent findings, the use of explainable AI tools is now considered essential under current regulatory pressure in cybersecurity and fraud detection domains \cite{jain2024}.

In this research, a stacking ensemble method is proposed and combined with a set of XAI tools. This combination offers a reliable and practical solution for fraud detection in real-world financial environments by merging high performance with interpretability.

\section{Literature Review}
This research focuses on three main areas: using advanced AI models for fraud detection, improving model performance through stacking ensemble methods, and making models more transparent using explainable artificial intelligence (XAI) techniques. The following section discusses related work in each of these areas and shows how they collectively help improve fraud detection in the financial field.

\subsection{Fraud Detection in Financial Transactions with AI Techniques}\label{sub1sec2}

It is essential to use reliable and flexible methods for fraud detection. Early research showed that ensemble models like Random Forest outperform traditional models such as SVM and logistic regression in fraud detection \cite{bhattacharyya2011}. Later studies addressed issues like class imbalance and suggested that models should be able to learn and adapt over time \cite{pozollo2014}.

Bagging techniques have been used to help reduce false positives \cite{zareapoor2015}. Unsupervised learning approaches were also explored. For example, one study proposed an adaptive feedback mechanism and an aggregation strategy to handle emerging fraud patterns \cite{pumsirirat2018}, \cite{jiang2018}. Meanwhile, deep learning methods such as autoencoders and Restricted Boltzmann Machines were applied to detect unusual activity \cite{pumsirirat2018}. AdaBoost-based models with majority voting achieved over 99\% accuracy \cite{randhawa2018}.

More recent research introduced hybrid deep learning methods and improved gradient boosting models. For example, LightGBM produced very good results when paired with strong feature engineering \cite{ge2020}. A hybrid model using binary cross-entropy and focal loss reached 95.8\% accuracy on an imbalanced dataset \cite{kewei2021}. A combined CatBoost and deep learning model showed promising outcomes, achieving AUC scores of up to 0.97 in certain experiments \cite{nguyen2022}. Deep autoencoders have also demonstrated high accuracy in credit card fraud detection, although performance varies across different datasets \cite{sharma2022}.

These studies clearly show that combining ensemble learning with deep learning techniques helps build strong and flexible fraud detection systems.

\subsection{Stacking Ensemble Methods in Fraud Detection }\label{sub2sec2}

Stacking ensemble methods have gained popularity for their ability to improve detection rates while simultaneously addressing class imbalance. Several new stacking models were introduced between 2023 and 2025. One design used SMOTE-ENN with LSTM and GRU networks, with an MLP as the meta-learner. It reached a perfect sensitivity (recall) of 1.00 and a specificity of 0.997 \cite{mienye2023}. Another study stacked an LSTM model and a Random Forest classifier, using an MLP as the meta-learner and applying SMOTE-ENN for class imbalance \cite{chellapilla2024}.

The CCAD model incorporated four outlier detectors and used XGBoost as a meta-learner. It also applied stratified sampling and a discordance learning technique to better catch rare fraud cases \cite{islam2023}. Another work combined XGBoost, CatBoost, and LightGBM in a stacking ensemble and employed Bayesian hyperparameter tuning to improve performance \cite{abdelghafour2024}.

Some models were designed to work with both financial and non-financial data. One such system integrated models from different domains and reached an AUC of 0.8146 with a recall of 0.8246, demonstrating utility in real-world applications \cite{zhu2024}.

In another study, a stacking model with a Random Forest as the meta-learner combined decision tree and Random Forest base learners. K-SMOTEENN was used to handle class imbalance, and LIME provided explanations for the results. This setup achieved an AUC of 1.00 and an F1-score of 0.92, outperforming even strong single models like XGBoost \cite{damanik2025}.

Stacking models can be highly effective for fraud detection, but they present several challenges. They are harder to understand, take longer to train, and risk overfitting, especially when many layers are used. However, if applied correctly – for example, with proper data balancing and XAI – they can be very powerful tools.

\subsection{Explainable AI (XAI) and Real-Time Scalability}\label{sub3sec2}

AI is being used more in fraud detection, and it is essential that the model can explain how it works. XAI tools help people understand what decisions the model is making and why.

Early tools like LIME and SHAP provided some of the first simple methods to explain black-box models \cite{ribeiro2016}, \cite{lundberg2017}. Later, SHAP was improved (e.g., Tree-SHAP) to work better with tree-based models, offering faster and more consistent explanations \cite{lundberg2018}.

At the same time, federated learning (FL) started to gain attention as a way to train models collaboratively without sharing raw data, thus preserving privacy. When combined with SMOTE, federated learning reached an AUC of 95.5\% on an imbalanced dataset \cite{yang2019}. One study showed that using XAI techniques with federated learning produced models that were both explainable and privacy-aware, achieving about 93\% accuracy \cite{awosika2024}.

Studies have explored the application of XAI in various domains, including cybersecurity \cite{zhang2022}. Some work has introduced guides to help practitioners choose the proper explanation method depending on the model \cite{martins2024}. Other researchers have focused on ensuring models are transparent and easy to audit in real time \cite{martins2024}.

In one study, a hybrid approach combining machine learning, deep learning, and five different XAI tools (including SHAP and LIME) achieved 98.3\% accuracy while remaining explainable \cite{sai2023}. Other papers have discussed ways to ensure models are fair, safe, and transparent by using ethical AI principles \cite{doshi-velez2017}.

Even though XAI helps explain a model’s decisions, it doesn’t solve every problem. Models may still exhibit bias or other issues. Overall, recent studies indicate that future fraud detection models must be accurate, explainable, scalable, and ethical in their design and deployment.

\section{Methodology}\label{sec3}

\begin{figure}[H]
\centering
\includegraphics[width=0.9\textwidth]{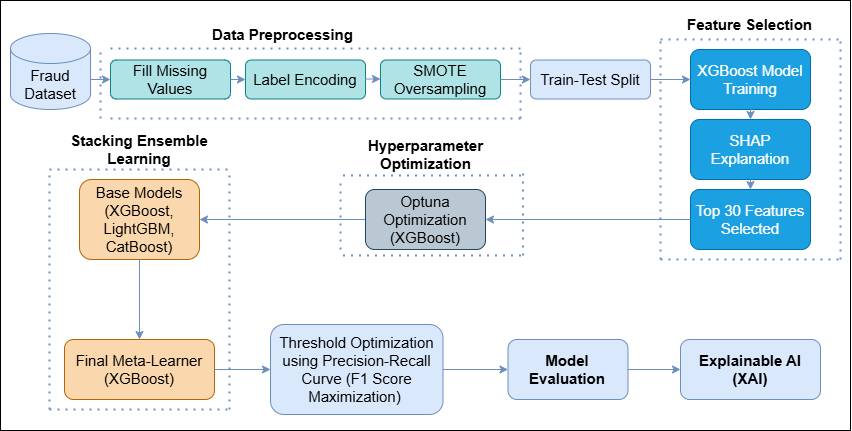}
\caption{The Proposed Fraud Detection Framework.}\label{fig1}
\end{figure}

This study introduces a robust framework for detecting financial fraud using XAI techniques. The project focuses on two primary objectives: developing a highly accurate model to identify fraudulent transactions and ensuring that its decisions are interpretable. The overall framework consists of several phases:

\begin{itemize}
\item Data collection
\item Data preparation
\item Feature selection
\item Hyperparameter optimization
\item Building a stacking ensemble model
\item Evaluation
\item XAI tools
\end{itemize}

Figure~\ref{fig1} presents the complete workflow of the proposed fraud detection system. Starting from data preprocessing and feature selection using SHAP, it shows the sequence of steps from model development through stacking ensemble learning, hyperparameter tuning with Optuna, and model evaluation. The final stage emphasizes how explainable AI methods are used to validate and interpret model predictions, thereby ensuring reliability and transparency.

\subsection{Data Collection}\label{sub1sec3}
This research utilizes the publicly available IEEE-CIS Fraud Detection dataset from a Kaggle competition \cite{ieeecis2025}. It encompasses nearly 590,000 transaction records and includes two types of data: transaction data and associated identity data.

\begin{table}[H]
\caption{Transaction Features.}\label{tab1}%
\begin{tabular}{@{}p{4cm}  p{8.3cm}@{}}
\toprule
Features & Description \\
\midrule
TransactionID    & A unique identifier is assigned to each transaction.   \\
TransactionDT    & Time delta from a reference datetime (timing information).   \\
TransactionAMT    & Transaction payment amount in USD.   \\
ProductCD    & Product code representing the product for the transaction.    \\
card1 – card6    & Payment card information (e.g., card type, issuing bank, country).   \\
addr1, addr2    & Address details linked to the transaction.   \\
dist1, dist2    & Distance metrics (e.g., distance between billing address and card address).    \\
P\_emaildomain, R\_emaildomain    & Email domains for purchaser and recipient.    \\
C1 – C14   & Count features (e.g., number of cards associated with address).    \\
D1 – D15    & Time deltas (e.g., days between transactions, or between user actions).   \\
M1 – M9    & Matching features (e.g., address and cardholder match flags).   \\
V1 – V339    & Vesta engineered features (ranking, counting, and entity relationships).    \\
isFraud    & Target variable: 1 for fraudulent transaction, 0 otherwise.    \\

\end{tabular}
\footnotetext{}

\end{table}

\begin{table}[H]
\caption{Identity Features.}\label{tab2}%
\begin{tabular}{@{}p{4cm}  p{8.3cm}@{}}
\toprule
Features & Description \\
\midrule
TransactionID   &  A unique identifier is assigned to each transaction. \\
 DeviceType  &   Type of device used for the transaction (e.g., desktop, mobile).\\
 DeviceInfo  &  Detailed device information (model, OS, etc.). \\
id\_01 – id\_11 &  Identity characteristics related to transaction (nature is masked). \\
  id\_12 – id\_38  &  Categorical identity features (network info, browser, proxy, IP address attributes, etc.). \\

\end{tabular}
\footnotetext{}

\end{table}

Table ~\ref{tab1} and Table~\ref{tab2} summarize the transaction and identity features, respectively, provided in the dataset. Key transaction features include transaction timestamps, amounts, product codes, card information, billing and email address data, and a large set of engineered features (V1–V339) generated by the data sponsor. The identity features (Table 2) cover device type and information, as well as various network attributes (masked IP, browser, etc.). Each transaction may have an associated identity record, and TransactionID serves as a common key linking the two tables. For analysis, the transaction and identity tables were merged on TransactionID, and this identifier was dropped afterward as it does not carry predictive information.

\subsection{Exploratory Data Analysis (EDA) }\label{sub2sec3}
We first examined various features to identify patterns distinguishing fraudulent and legitimate transactions. For example: 
\begin{figure}[H]
\centering
\includegraphics[width=0.9\textwidth]{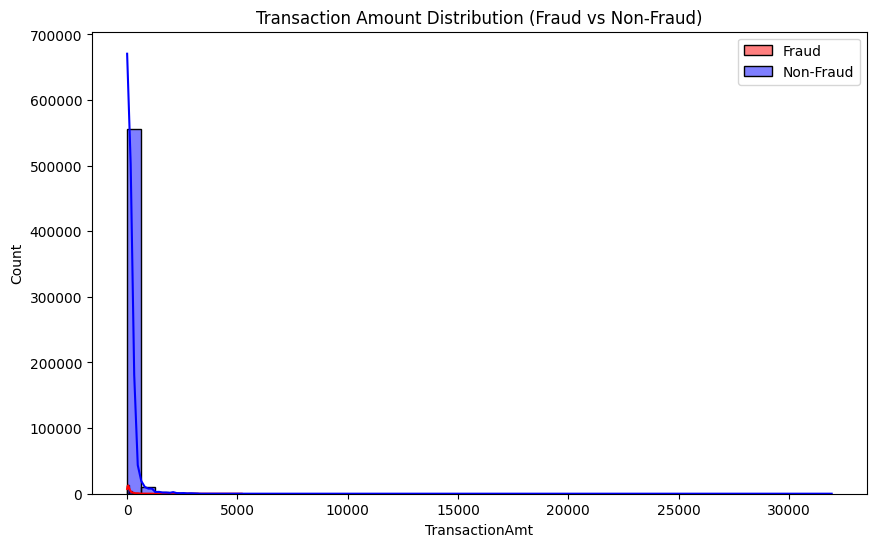}
\caption{The Transaction Amount distributions.}\label{fig2}
\end{figure}
Transaction Amount (Figure~\ref{fig2}) is highly right-skewed; most fraudulent transactions involve relatively small amounts (below \$1000).

\begin{figure}[H]
\centering
\includegraphics[width=0.9\textwidth]{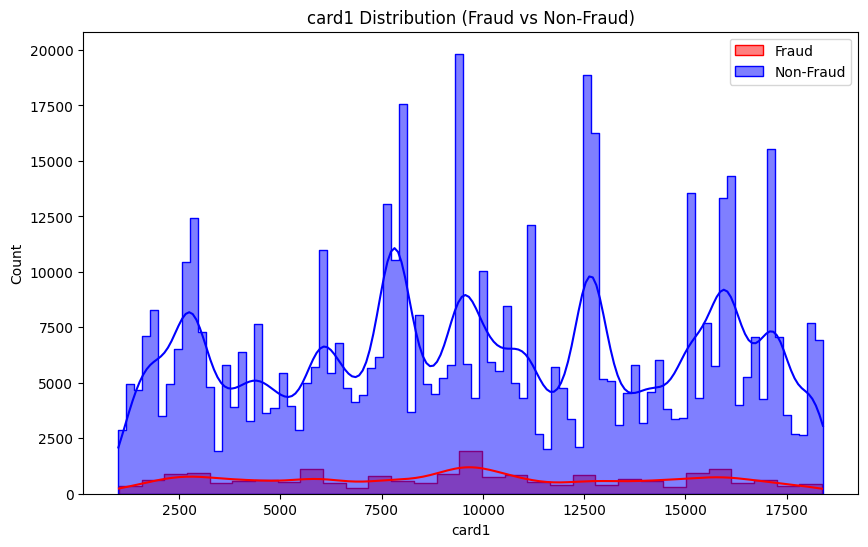}
\caption{Transaction Amount distributions.  }\label{fig3}
\end{figure}
The card1 feature (Figure~\ref{fig3}) showed distinct clustering for non-fraudulent transactions, whereas fraudulent transaction values were more dispersed. 

\begin{figure}[H]
\centering
\includegraphics[width=0.9\textwidth]{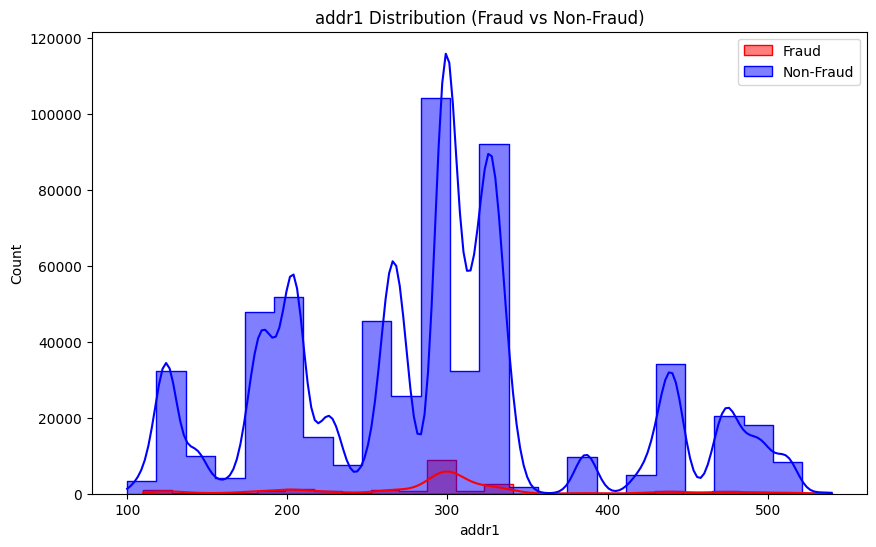}
\caption{addr1 distributions. 
 }\label{fig4}
\end{figure}

Similarly, addr1 (Figure~\ref{fig4}) exhibited notable peaks for non-fraud cases, while fraud cases were more evenly spread across address values.

Many features in the dataset have obfuscated or unclear meanings (e.g., the engineered V-features), which limits the value of direct visualization. With more than 590,000 transactions and hundreds of features, a purely manual EDA is challenging. Instead, we utilized SHAP-based analysis for feature importance to quickly and accurately identify the most impactful features for fraud detection.

\subsection{Data Preprocessing }\label{sub3sec3}
\begin{figure}[H]
\centering
\includegraphics[width=0.9\textwidth]{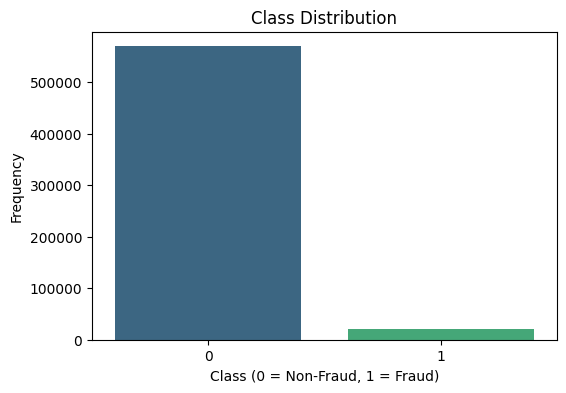}
\caption{Class Distribution Before SMOTE. 
 }\label{fig5}
\end{figure}

\begin{figure}[H]
\centering
\includegraphics[width=0.9\textwidth]{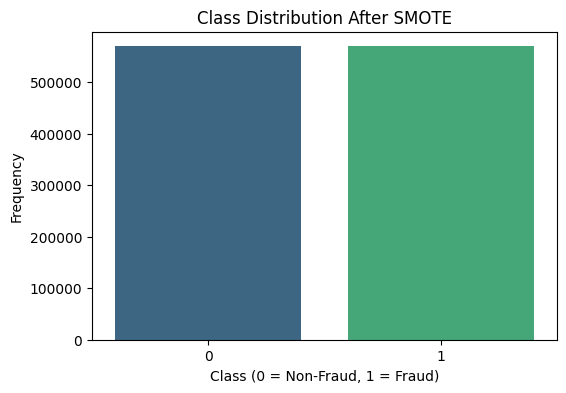}
\caption{Class Distribution After SMOTE. 
 }\label{fig6}
\end{figure}

To prepare the data for modeling, we performed several preprocessing steps:

\begin{enumerate}
\item Removing unnecessary identifiers: Unique identifiers such as TransactionID were dropped, as they do not contribute to fraud detection. Merging the transaction and identity tables made TransactionID redundant. 

\item Handling missing data: Missing values can degrade model performance. We imputed missing categorical values with the most frequent category, and missing numerical values with the median of each feature. This approach preserves the distribution of each feature while mitigating the impact of outliers on imputation.

\item Encoding categorical variables: Categorical features were converted to numeric representations using label encoding. This ensured that the machine learning models could process these features.

\item Addressing class imbalance: Fraud detection datasets are typically highly imbalanced. In our dataset, only about 3.5\% of transactions were fraudulent, with the vast majority (96.5\%) being legitimate. This imbalance can bias the model toward predicting the majority class (non-fraud). To balance the training data, we applied the Synthetic Minority Over-sampling Technique (SMOTE), which generates synthetic minority-class examples. After applying SMOTE, the training set had an approximately equal distribution of fraud and non-fraud samples (Figure~\ref{fig5} and Figure~\ref{fig6}). We then split the dataset into training and testing subsets (80\% train, 20\% test), ensuring that the model’s performance would be evaluated on unseen (imbalanced) test data.
\end{enumerate}

\subsection{Feature Selection }\label{sub4sec3}

To optimize model performance and reduce dimensionality, feature selection was driven by explainability techniques. We first trained an XGBoost classifier on the full training data (with all features). We then computed SHAP values for all features to assess their importance. The top 30 features with the highest mean absolute SHAP values were selected for modeling.

\begin{figure}[H]
\centering
\includegraphics[width=0.9\textwidth]{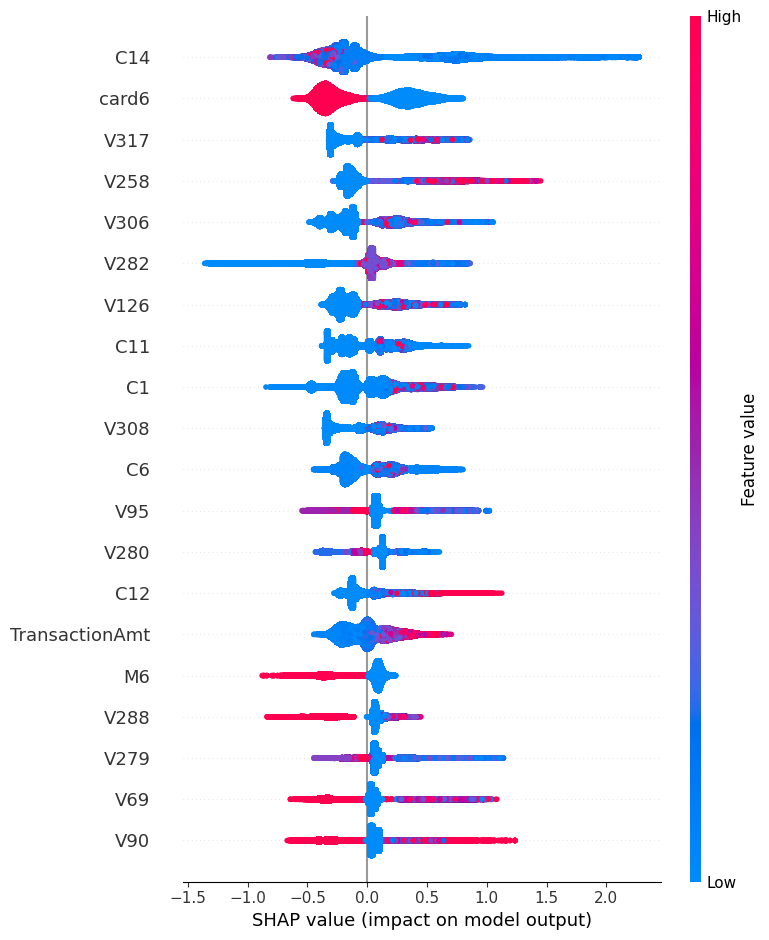}
\caption{SHAP Summary Plot. 
 }\label{fig7}
\end{figure}

This SHAP-based approach helped pinpoint the most influential features in determining fraud, which streamlines the model and potentially improves generalization. Figure~\ref{fig7} (SHAP summary plot) highlights the features that strongly influence the model’s predictions: each feature is sorted by importance, and the distribution of SHAP values indicates how variations in that feature drive the model output.

\subsection{Hyperparameter Optimization  }\label{subs5ec3}

Hyperparameters are the settings that govern the learning process of the model (for example, the learning rate or maximum tree depth in a booster). We employed Optuna for automated hyperparameter tuning. Optuna performed 20 trial runs to find the best hyperparameter values for our XGBoost model. The key hyperparameters tuned included: 

\begin{itemize}
\item n\_estimators: Number of trees in the model.
\item max\_depth: Maximum depth of each tree.
\item learning\_rate: How quickly the model learns (trade-off between speed and convergence).
\item subsample: Fraction of training samples used for each tree.
\item colsample\_bytree: Fraction of features used when building each tree.
\item scale\_pos\_weight: Balancing weight to assign to the positive (fraud) class due to class imbalance.

\end{itemize}
 
After tuning, the XGBoost model was retrained using the optimal hyperparameters that Optuna identified.

\subsection{Model Development: Stacking Ensemble}\label{sub6sec3}

We used a stacking ensemble method instead of relying on a single classifier. In a stacking ensemble, multiple base models are combined, and a final model (meta-learner) learns how to best blend their outputs.
Our ensemble included three high-performing gradient boosting models as base learners:

\begin{itemize}
\item XGBoost: Known for its efficiency on large datasets and ability to capture complex patterns. 
\item LightGBM: A fast, high-performance booster that can handle a large number of features efficiently.
\item CatBoost: A gradient boosting model particularly effective for handling categorical features.

\end{itemize}

Each of the three base models was trained on the training data (using the top 30 features). For each instance, the base models produced predictions, which were then used as input features for the meta-learner. We chose a simple XGBoost classifier as the meta-learner. This meta-learner was trained to take the three base model predictions and output the final fraud/non-fraud decision. This layered design allows the meta-model to capture relationships or complementarities that individual models might overlook.

\subsection{Cross-Validation }\label{sub7sec3}

To ensure the model generalizes well and to mitigate overfitting, we used 5-fold stratified cross-validation during training. Stratification preserved the fraud/non-fraud class ratio in each fold, which is important given the class imbalance. In each iteration, the model was trained on 80\% of the data (with SMOTE applied to the training portion) and validated on the remaining 20\% (with original class distribution). We tracked performance metrics (primarily AUC) on each fold. The results were then averaged across the five folds to assess the model’s stability.

\subsection{Evaluation Metrics }\label{sub8sec3}
\begin{itemize}
\item \textbf{Precision}: The proportion of transactions flagged as fraud that were actually fraudulent (i.e., positive predictive value). 
\item \textbf{Recall}: The proportion of actual fraud cases that the model correctly detected (sensitivity).
\item \textbf{F1-Score}: The harmonic mean of precision and recall, providing a single measure that balances both.
\item \textbf{AUC-ROC}: The Area Under the ROC Curve, which indicates how well the model separates fraud vs. non-fraud across all classification thresholds. 
\item \textbf{Confusion Matrix}: A summary of correct and incorrect predictions, broken down by class (fraud or not fraud).
\item \textbf{Threshold Optimization}: We also examined different probability thresholds for classifying a transaction as fraud, selecting the threshold that maximized the F1-score on the validation data.

\end{itemize}

\subsection{Explainable Artificial Intelligence (XAI)}\label{sub3sec9}

We applied several post-hoc explainability methods to interpret the stacking ensemble’s predictions:
\begin{itemize}
\item \textbf{SHAP}: Used to compute feature importance and perform feature selection by identifying the top 30 features contributing most to the model output.
\item \textbf{LIME}: We utilized LIME to explain individual predictions, by locally approximating the model with an interpretable model to see which features drive a specific prediction.
\item \textbf{Partial Dependence Plots (PDPs}: Show how the model’s predicted fraud probability changes when we vary one feature at a time while holding others constant.
\item \textbf{Permutation Feature Importance}: Involves randomly shuffling the values of each feature and measuring the decrease in model performance, thereby indicating the feature’s importance.

\end{itemize}

\section{Results}\label{sec4}

When evaluating the model, we aimed to achieve the highest accuracy while also ensuring its decisions can be clearly understood and trusted.

\subsection{Model Evaluation Metrics }\label{sub1sec4}

\begin{figure}[H]
\centering
\includegraphics[width=0.9\textwidth]{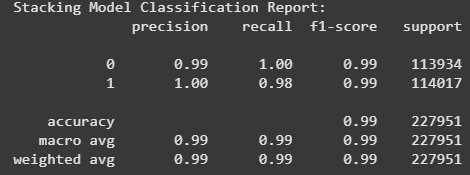}
\caption{Classification Report. 
 }\label{fig8}
\end{figure}

Using the top 30 SHAP-selected features, we trained our stacking ensemble. Training took approximately 278 seconds, reflecting the model’s efficiency despite the large dataset. The performance metrics on the test set are summarized in the classification report (Figure~\ref{fig8}).

The model identified almost all legitimate (non-fraud) transactions with few errors, achieving a recall of 1.00 and a precision of 0.99 for the non-fraud class. It was similarly strong for the fraud class, with a precision of 1.00 and a recall of 0.98. Averaged across both classes, the precision, recall, and F1-score were all about 0.99, indicating balanced and excellent performance. With 99\% overall accuracy, the model appears well-suited for real-world fraud detection scenarios where both precision and recall are critical.

\subsection{Cross-Validation Results }\label{sub2sec4}

We further verified the model’s robustness through 5-fold cross-validation (using ROC-AUC as the evaluation metric). The AUC scores for the five folds were: [0.9979, 0.9979, 0.9979, 0.9981, 0.9980]. These scores were very consistent (all approximately 0.998), demonstrating that the model’s performance is stable across different subsets of the data and not overly sensitive to a particular train-test split.

\subsection{Confusion Matrix  }\label{sub3sec4}
\begin{figure}[H]
\centering
\includegraphics[width=0.9\textwidth]{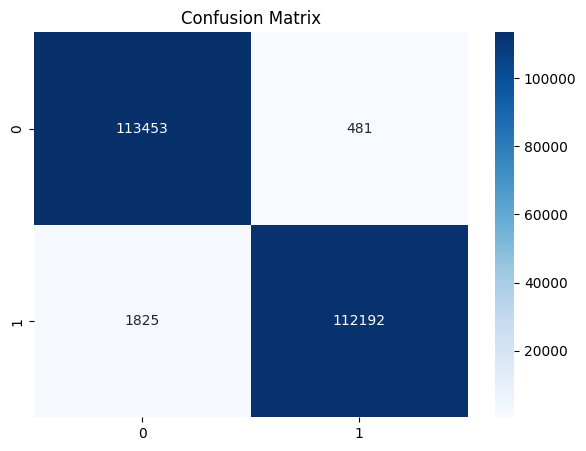}
\caption{Confusion Matrix. 
 }\label{fig9}
\end{figure}

Figure~\ref{fig9} shows the confusion matrix of the model’s predictions on the Training set.

Figure~\ref{fig9} shows the confusion matrix of the model’s predictions on the test set (with the original class imbalance). The breakdown is as follows:

\begin{itemize}
\item True Negatives (113,453): Legitimate transactions correctly identified as non-fraudulent. 
\item False positives (481): Legitimate transactions that were mistakenly flagged as fraudulent.
\item False Negatives (1,825): Fraudulent transactions that the model failed to detect (incorrectly classified as non-fraud).
\item True Positives (112,192): Fraudulent transactions correctly detected by the model.

\end{itemize}

These values correspond to a recall of ~100\% for non-fraud (almost no legitimate transaction was misclassified) and ~98\% recall for fraud, with precision ~99\% for both classes (as noted above). The extremely low false positive rate (481 out of ~113,934 legitimate) and low false negative rate (1,825 out of ~114,017 fraud) illustrate the model’s strong performance in both detecting fraud and avoiding false alarms.

\subsection{Curve and Precision-Recall Curve }\label{sub4sec4}

\begin{figure}[H]
\centering
\includegraphics[width=0.9\textwidth]{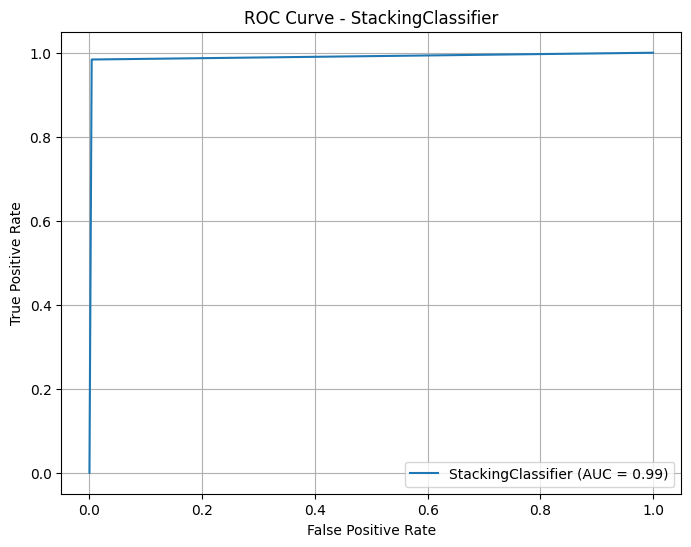}
\caption{ROC Curve of Stacking Classifier.
 }\label{fig10}
\end{figure}

The ROC curve for our stacking classifier (Figure~\ref{fig10}) is very close to the top-left corner of the plot, and the AUC is 0.99. A curve that hugs the top-left indicates near-perfect classification performance; our model’s ROC confirms its outstanding ability to separate fraud vs. non-fraud cases.

\begin{figure}[H]
\centering
\includegraphics[width=0.9\textwidth]{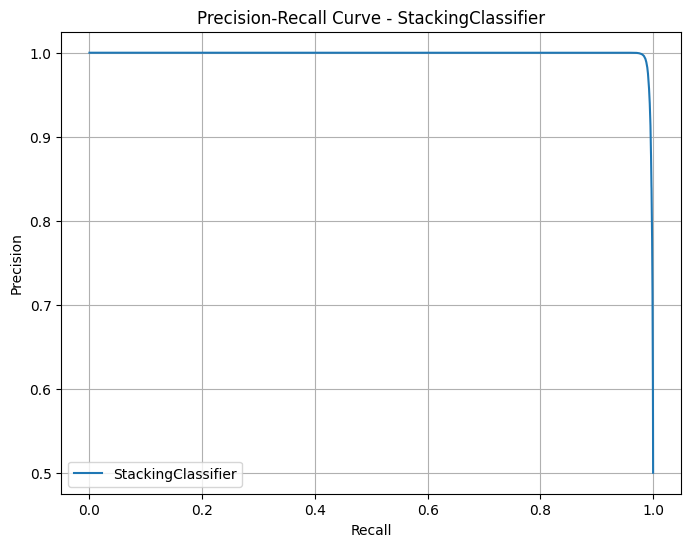}
\caption{Precision-Recall (PR) Curve of Stacking Classifier. 
 }\label{fig11}
\end{figure}
The PR curve (Figure~\ref{fig11}) highlights the model’s ability to detect fraud in the context of class imbalance. The curve shows that the model maintains very high precision even as recall increases, meaning it catches the majority of fraud cases with very few false alarms. This is especially important because fraudulent transactions are rare relative to normal ones; a high precision ensures that investigators’ attention is not wasted on too many false fraud alerts.

\subsection{Threshold Optimization }\label{sub5sec4}

We used the precision-recall curve to find the probability threshold that maximizes the F1-score. The optimal threshold was about 0.44, slightly below the default 0.5 cutoff. Using this lower threshold improved the balance between precision and recall for the fraud class. In practice, this means the model was tuned to catch a few extra fraud cases (improving recall) at the cost of a very small increase in false positives, thus maximizing overall fraud detection effectiveness as measured by F1-score.

\subsection{Explainable Artificial Intelligence (XAI) Results  }\label{sub6sec4}

\subsubsection{LIME}\label{sub1sub6sec4}

\begin{figure}[H]
\centering
\includegraphics[width=0.9\textwidth]{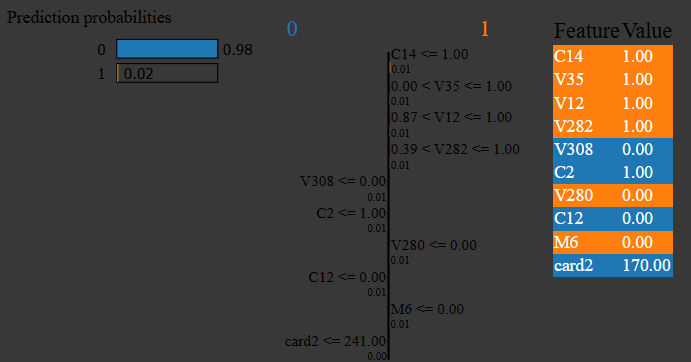}
\caption{LIME Explanation Example.
 }\label{fig12}
\end{figure}

Understanding the rationale behind a specific prediction can be challenging for an intricate model such as a stacking ensemble. LIME addresses this by constructing a simplified local model around an individual prediction, making it possible to identify which features most significantly influenced that particular decision. For example, in one transaction’s explanation (Figure~\ref{fig12}), features shown on the right side pushed the prediction toward fraud (orange bars) or toward non-fraud (blue bars). In that case, features “C14” and “V12” strongly pushed the model toward a non-fraud decision, resulting in a 98\% predicted probability for the non-fraud class. Such local explanations increase trust in the model by showing plausible reasons for each automated decision.

\subsubsection{Permutation Feature Importance}\label{sub2sub6sec4}

\begin{figure}[H]
\centering
\includegraphics[width=0.9\textwidth]{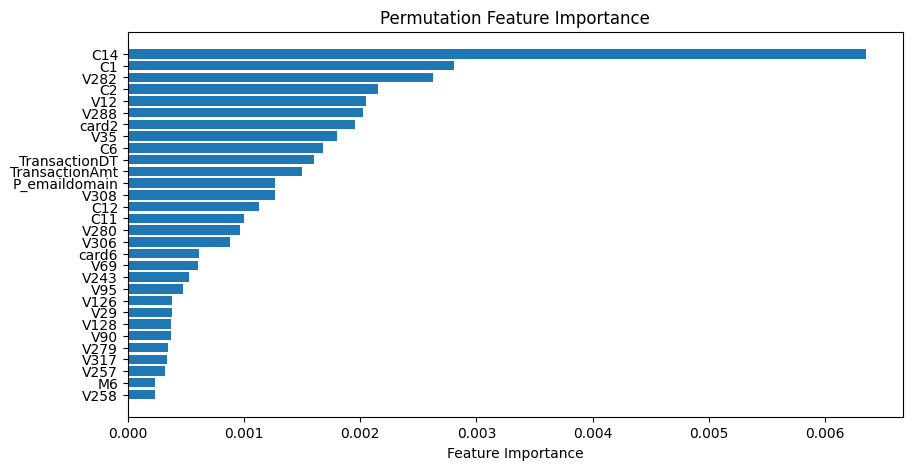}
\caption{Permutation Feature Importance.
 }\label{fig13}
\end{figure}

This technique involves shuffling the values of each feature and observing how much the model’s performance degrades as a result. In our analysis, we calculated permutation importance for the trained stacking model (Figure~\ref{fig13}). The resulting plot shows that features with longer bars (greater performance drop when shuffled) are more important. For instance, “C14” had the longest bar in Figure~\ref{fig13}, suggesting it is the most influential feature for detecting fraud in our model. Other top features follow in order. This aligns with the earlier SHAP findings that “C14” (and some V-features) carry significant predictive power.

\subsubsection{Permutation Feature Importance}\label{sub3sub6sec4}

\begin{figure}[H]
\centering
\includegraphics[width=0.9\textwidth]{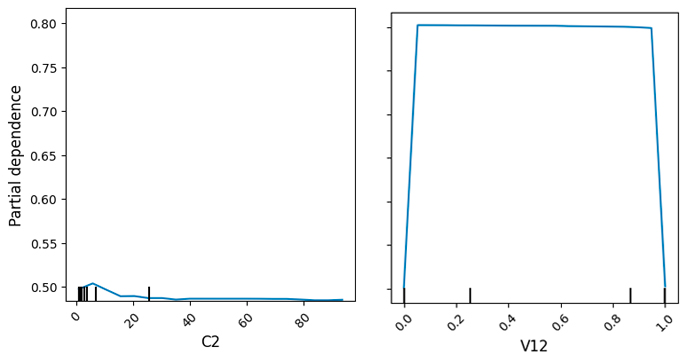}
\caption{Partial Dependence Plots.
 }\label{fig14}
\end{figure}

PDPs illustrate how the model’s predicted fraud probability changes when we vary a single feature while keeping all others constant. This is akin to turning one input “dial” to see its effect on the output. Figure~\ref{fig14} shows PDPs for selected important features. For example, for feature “V12”, when its value is around 0.5 (some normalized scale), the model is more likely to predict fraud. In contrast, the PDP for “C2” indicates that as “C2” increases, the predicted fraud probability tends to remain flat or even decrease. Such insights help in understanding the global behavior of the model with respect to key features – for instance, certain features might have non-linear or threshold effects on fraud risk.

\subsection{Comparative Analysis }\label{sub7sec4}

We evaluated the proposed stacking ensemble by comparing it against some traditional classifiers and recent advanced approaches. Table~\ref{tab3} compares our model’s performance to that of a Logistic Regression and a single Decision Tree classifier (both trained on the same features for a fair comparison). Our Proposed Model achieved 0.99 accuracy and an AUC of 0.9983, with precision/recall/F1 of 0.98/1.00/0.99 for the non-fraud class and 1.00/0.98/0.99 for the fraud class. In contrast, the Logistic Regression achieved only 0.61 accuracy (AUC 0.6614), with considerably lower recall for fraud (50\%). The Decision Tree performed better than Logistic Regression (0.87 accuracy, AUC 0.9250), but still significantly underperformed our ensemble, especially in precision-recall balance. These results illustrate that our stacking ensemble outperforms baseline models by a wide margin, particularly in handling the class imbalance (as seen by the much higher recall for the minority fraud class).

\begin{table}[H]
\caption{Comparison Table: Traditional Models.}\label{tab3}%
\small 
\begin{tabular}{@{}p{4cm} p{2cm}p{2cm} p{4.3cm}@{}}
\toprule
Model & Accuracy & AUC  & Precision / Recall / F1-Score \\
\midrule
 Our
Proposed Model
 & 0.99  & 0.9983 &  0.98 / 1.00 / 0.99 (Class 0) 
1.00 / 0.98 / 0.99 (Class 1)
 \\
 Logistic Regression & 0.61  & 0.6614 &  0.59 / 0.72 / 0.65 (Class 0)
0.65 / 0.50 / 0.57 (Class 1)
 \\
  Decision Tree   & 0.87  & 0.9250 &  0.88 / 0.85 / 0.87 (Class 0)
0.86 / 0.88 / 0.87 (Class 1)
\\

\end{tabular}
\footnotetext{}

\end{table}

We also compare our work with several recent studies (Table~\ref{tab4}) to contextualize the performance and approach. Our proposed model (stacking XGBoost, LightGBM, CatBoost with tuned hyperparameters) achieved 99\% accuracy, 0.99 precision, 0.99 recall, 0.99 F1, and 0.99 AUC on a large real-world dataset. In comparison:

\begin{table}[H]
\caption{Comparison Table: Studies Models.}\label{tab4}%
\small 
\begin{tabular}{@{}p{2.5cm} p{6.5cm} p{5cm}@{}}
\toprule
Reference & Model Description  & Accuracy  \\
\midrule
 Our
Proposed Model
 & Stacking (XGBoost + LightGBM + CatBoost with optimized hyperparameters)& Accuracy: 0.99, Precision: 0.99, Recall: 0.99, F1-Score: 0.99, AUC: 0.99
 \\
Ref \cite{sai2023} & LightGBM (Explainable AI approach, tuned)  &  Accuracy 98.3\%, F1-score 0.70, AUC-ROC 0.96
 \\
 Ref \cite{zhu2024}  &  Stacking Ensemble (Random Forest + AdaBoost + XGBoost + LightGBM + ExtraTrees with Logistic Regression meta-learner)  &   Recall: 0.8246, AUC: 0.8146 
\\
 Ref \cite{damanik2025}  & Stacking (Random Forest + Decision Tree with Random Forest meta-learner + K-SMOTEENN + feature construction)   &   F1: 0.92, Precision: 0.95, Recall: 0.88, ROC-AUC: 1.00
\\
 Ref \cite{veigas2021}  & Optimized Stacking Ensemble (MLP + kNN + SVM with Logistic Regression meta-learner using SMOTE+GAN)   &   Accuracy 99.86\%, F1 improvement 16\%
\\
Ref \cite{talukder2024}   &  Hybrid Ensemble (DT + RF + KNN + MLP, Grid Search Optimization, IHT-LR balancing)  &   Accuracy 99.66–100\% depending on the model 
\\

\end{tabular}
\end{table}

Compared to these studies, our framework’s results are similar or even better than the state-of-the-art, especially considering we evaluated on a large, real-world dataset. While some studies show near-perfect metrics, those often involve either very small or heavily synthesized datasets (as noted by their methodology). By contrast, our model performed exceedingly well on a large and complex real-world dataset, underscoring the practicality and robustness of our approach.

\section{Discussion and Conclusion}\label{sec5}

This research confirms the growing significance of XAI in financial fraud detection. Machine learning algorithms have shown high accuracy in detecting fraudulent activity (e.g., ensemble models in \cite{bhattacharyya2011}, \cite{randhawa2018}, but they often lack interpretability (highlighted by concerns in \cite{deloitte2024}, \cite{gunning2019}). Our approach combines a stacking ensemble with model interpretation tools such as SHAP and LIME to address this gap.

Recent work shows a growing reliance on ensemble techniques, particularly stacking, for their ability to integrate multiple models and reduce overfitting \cite{chellapilla2024}, \cite{abdelghafour2024}, \cite{zhu2024}. Reflecting that trend, our model delivered strong results: both precision and recall reached 0.99, leading to an equally high F1-score, and an AUC-ROC of 0.99. These metrics are on par with top-performing models reported in the literature (for instance, our results align with the performance of the stacking model in \cite{damanik2025}).

Methods such as SHAP and LIME have provided essential insights into feature contributions at global and local levels, emphasizing the necessity for explainable models in financial AI systems \cite{lundberg2017}, \cite{ribeiro2016}. Our use of SHAP for feature selection and LIME for individual prediction explanations builds on this prior work and demonstrates how interpretability techniques can be seamlessly integrated into a high-performing fraud detection framework \cite{gunning2019}, \cite{zhang2022}, \cite{martins2024}.

The proposed approach demonstrated higher performance compared to traditional models (like logistic regression or single decision trees), as well as against models that use data balancing techniques like K-SMOTEENN or SMOTE-ENN \cite{mienye2023}, \cite{chellapilla2024}, \cite{abdelghafour2024}. While some existing models report slightly better metrics on synthetic or limited datasets, our study underscores real-world relevance by using a large-scale, highly imbalanced dataset \cite{ieeecis2025}.

The developed framework, which integrates XGBoost, LightGBM, and CatBoost with interpretation techniques like SHAP, LIME, and Permutation Importance, effectively addresses key challenges in modern fraud detection, such as model transparency and data imbalance \cite{damanik2025}, \cite{ieeecis2025}. It outperformed many conventional and advanced models, showing that high accuracy can be achieved without turning models into “black boxes” \cite{deloitte2024}, \cite{gunning2019}.

Ensuring that AI systems are both accurate and explainable is vital for user trust, regulatory compliance, and responsible AI deployment, especially in finance \cite{jain2024}, \cite{martins2024}. Future research may explore real-time deployment challenges and solutions (e.g., optimizing the speed of explanations for instant fraud alerts), adapting models to evolving fraud patterns through continual learning, and evaluating performance within privacy-preserving frameworks such as federated learning \cite{yang2019}, \cite{awosika2024}, \cite{yang2019federated}. Incorporating additional aspects like fairness and adversarial robustness into fraud detection models would also be worthwhile, to ensure these systems remain reliable and unbiased as they are widely adopted.

As digital financial systems continue to grow in scale and complexity, developing explainable and ethical AI solutions will be essential to maintain transparency, trust, and security. The results of this study provide evidence that it is feasible to combine cutting-edge machine learning techniques with XAI to create fraud detection models that are not only highly accurate but also interpretable and trustworthy in practice.


\begin{thebibliography}{99}

\bibitem{ibm2024}
IBM Newsroom. *Data Suggests Growth in Enterprise Adoption of AI is Due to Widespread Deployment by Early Adopters*. 2024. Accessed: April 23, 2025. Available:\url{https://newsroom.ibm.com/2024-01-10-Data-Suggests-Growth-in-Enterprise-Adoption-of-AI-is-Due-to-Widespread-Deployment-by-Early-Adopters}



\bibitem{deloitte2024}
Deloitte Insights. *Unleashing the power of machine learning models in banking through explainable artificial intelligence (XAI)*. 2024. Accessed: April 23, 2025. Available: \url{https://www2.deloitte.com/us/en/insights/industry/financial-services/explainable-ai-in-banking.html}

\bibitem{arrieta2020}
Barredo Arrieta, A. et al. *Explainable Artificial Intelligence (XAI): Concepts, taxonomies, opportunities and challenges toward responsible AI*. Information Fusion, vol. 58, pp. 82--115, 2020. doi: 10.1016/j.inffus.2019.12.012

\bibitem{jain2024}
Jain, V., Balakrishnan, A., Beeram, D., Najana, M., and Chintale, P. *Leveraging Artificial Intelligence for Enhancing Regulatory Compliance in the Financial Sector*. Int. J. Comput. Trends Technol., vol. 72, no. 5, pp. 124--140, May 2024. doi: 10.14445/22312803/IJCTT-V72I5P116

\bibitem{gunning2019}
Gunning, D., Stefik, M., Choi, J., Miller, T., Stumpf, S., and Yang, G.-Z. *XAI—Explainable artificial intelligence*. Science Robotics, vol. 4, no. 37, p. eaay7120, Dec. 2019. doi: 10.1126/scirobotics.aay7120

\bibitem{webb2025}
Webb, M. *60+ Global Credit Card Fraud Statistics in 2025*. 2025. Accessed: April 23, 2025. Available: \url{https://www.techopedia.com/credit-card-fraud-statistics}

\bibitem{veigas2021}
Veigas, K. C., Regulagadda, D. S., and Kokatnoor, S. A. *Optimized Stacking Ensemble (OSE) for Credit Card Fraud Detection using Synthetic Minority Oversampling Model*. Indian J. Sci. Technol., vol. 14, no. 32, pp. 2607--2615, Aug. 2021. doi: 10.17485/IJST/v14i32.807

\bibitem{bhattacharyya2011}
Bhattacharyya, S., Jha, S., Tharakunnel, K., and Westland, J. C. *Data mining for credit card fraud: A comparative study*. Decision Support Systems, vol. 50, no. 3, pp. 602--613, Feb. 2011. doi: 10.1016/j.dss.2010.08.008

\bibitem{pozollo2014}
Dal Pozzolo, A., Caelen, O., Le Borgne, Y.-A., Waterschoot, S., and Bontempi, G. *Learned lessons in credit card fraud detection from a practitioner perspective*. Expert Syst. Appl., vol. 41, no. 10, pp. 4915--4928, Aug. 2014. doi: 10.1016/j.eswa.2014.02.026

\bibitem{zareapoor2015}
Zareapoor, M., and Shamsolmoali, P. *Application of Credit Card Fraud Detection: Based on Bagging Ensemble Classifier*. Procedia Comput. Sci., vol. 48, pp. 679--685, 2015. doi: 10.1016/j.procs.2015.04.201

\bibitem{pumsirirat2018}
Pumsirirat, A., and Yan, L. *Credit Card Fraud Detection using Deep Learning based on Auto-Encoder and Restricted Boltzmann Machine*. Int. J. Adv. Comput. Sci. Appl., vol. 9, no. 1, 2018. doi: 10.14569/IJACSA.2018.090103


\bibitem{jiang2018}
Jiang, C., Song, J., Liu, G., Zheng, L., and Luan, W. *Credit Card Fraud Detection: A Novel Approach Using Aggregation Strategy and Feedback Mechanism*. IEEE Internet Things J., vol. 5, no. 5, pp. 3637--3647, Oct. 2018. doi: 10.1109/JIOT.2018.2816007

\bibitem{randhawa2018}
Randhawa, K., Loo, C. K., Seera, M., Lim, C. P., and Nandi, A. K. *Credit Card Fraud Detection Using AdaBoost and Majority Voting*. IEEE Access, vol. 6, pp. 14277--14284, 2018. doi: 10.1109/ACCESS.2018.2806420

\bibitem{ge2020}
Ge, D., Gu, J., Chang, S., and Cai, J. *Credit Card Fraud Detection Using LightGBM Model*. Proc. 2020 Int. Conf. E-Commerce Internet Technol. (ECIT), pp. 232--236, Zhangjiajie, China, Apr. 2020. doi: 10.1109/ECIT50008.2020.00060

\bibitem{kewei2021}
Kewei, X., Peng, B., Jiang, Y., and Lu, T. *A Hybrid Deep Learning Model For Online Fraud Detection*. Proc. 2021 IEEE Int. Conf. Consumer Electron. Comput. Eng. (ICCECE), pp. 431--434, Guangzhou, China, Jan. 2021. doi: 10.1109/ICCECE51280.2021.9342110

\bibitem{nguyen2022}
Nguyen, N. et al. *A Proposed Model for Card Fraud Detection Based on CatBoost and Deep Neural Network*. IEEE Access, vol. 10, pp. 96852--96861, 2022. doi: 10.1109/ACCESS.2022.3205416

\bibitem{sharma2022}
Sharma, M. A., Raj, B. R. G., Ramamurthy, B., and Bhaskar, R. H. *Credit Card Fraud Detection Using Deep Learning Based on Auto-Encoder*. ITM Web Conf., vol. 50, p. 01001, 2022. doi: 10.1051/itmconf/20225001001

\bibitem{mienye2023}
Mienye, I. D., and Sun, Y. *A Deep Learning Ensemble With Data Resampling for Credit Card Fraud Detection*. IEEE Access, vol. 11, pp. 30628--30638, 2023. doi: 10.1109/ACCESS.2023.3262020

\bibitem{chellapilla2024}
Chellapilla, V., Chikkam, S., Melinati, J. S., and Ezhilarasan, M. *Credit Card Fraud Detection Using a Stacking Ensemble Approach With LSTM and Random Forest Machine Learning Techniques*. Int. Educ. Res. J., vol. 10, no. 4, Apr. 2024. doi: 10.21276/IERJ24540778981227

\bibitem{islam2023}
Islam, M. A., Uddin, M. A., Aryal, S., and Stea, G. *An Ensemble Learning Approach for Anomaly Detection in Credit Card Data with Imbalanced and Overlapped Classes*. J. Inf. Secur. Appl., vol. 78, p. 103618, Nov. 2023. doi: 10.1016/j.jisa.2023.103618

\bibitem{abdelghafour2024}
Abdelghafour, E. B., Mohamed, C., Noura, A., and Abdelhamid, B. *Enhancing Credit Card Fraud Detection Using a Stacking Model Approach and Hyperparameter Optimization*. Int. J. Adv. Comput. Sci. Appl., vol. 15, no. 10, 2024. doi: 10.14569/IJACSA.2024.01510110



\bibitem{zhu2024}
Zhu, S. et al. *A Financial Fraud Prediction Framework Based on Stacking Ensemble Learning*. Systems, vol. 12, no. 12, p. 588, Dec. 2024. doi: 10.3390/systems12120588

\bibitem{damanik2025}
Damanik, N., and Liu, C.-M. *Advanced Fraud Detection: Leveraging K-SMOTEENN and Stacking Ensemble to Tackle Data Imbalance and Extract Insights*. IEEE Access, vol. 13, pp. 10356--10370, 2025. doi: 10.1109/ACCESS.2025.3528079

\bibitem{ribeiro2016}
Ribeiro, M. T., Singh, S., and Guestrin, C. *‘Why Should I Trust You?’: Explaining the Predictions of Any Classifier*. Proc. 22nd ACM SIGKDD Int. Conf. Knowledge Discovery Data Mining, pp. 1135--1144, San Francisco, 2016. doi: 10.1145/2939672.2939778

\bibitem{lundberg2017}
Lundberg, S. M., and Lee, S.-I. *A Unified Approach to Interpreting Model Predictions*. Adv. Neural Inf. Process. Syst., 2017. Accessed: April 23, 2025. Available: \url{https://proceedings.neurips.cc/paper_files/paper/2017/hash/8a20a8621978632d76c43dfd28b67767-Abstract.html}

\bibitem{lundberg2018}
Lundberg, S. M., Erion, G. G., and Lee, S.-I. *Consistent Individualized Feature Attribution for Tree Ensembles*. arXiv preprint, 2018. doi: 10.48550/ARXIV.1802.03888

\bibitem{yang2019}
Yang, W., Zhang, Y., Ye, K., Li, L., and Xu, C.-Z. *FFD: A Federated Learning Based Method for Credit Card Fraud Detection*. In: Chen, K., Seshadri, S., Zhang, L.-J. (eds) Big Data – BigData 2019. Lecture Notes in Computer Science, vol. 11514. Springer, Cham, pp. 18--32. doi: 10.1007/978-3-030-23551-2\_2

\bibitem{awosika2024}
Awosika, T., Shukla, R. M., and Pranggono, B. *Transparency and Privacy: The Role of Explainable AI and Federated Learning in Financial Fraud Detection*. IEEE Access, vol. 12, pp. 64551--64560, 2024. doi: 10.1109/ACCESS.2024.3394528

\bibitem{zhang2022}
Zhang, Z., Hamadi, H. A., Damiani, E., Yeun, C. Y., and Taher, F. *Explainable Artificial Intelligence Applications in Cyber Security: State-of-the-Art in Research*. IEEE Access, vol. 10, pp. 93104--93139, 2022. doi: 10.1109/ACCESS.2022.3204051

\bibitem{martins2024}
Martins, T., De Almeida, A. M., Cardoso, E., and Nunes, L. *Explainable Artificial Intelligence (XAI): A Systematic Literature Review on Taxonomies and Applications in Finance*. IEEE Access, vol. 12, pp. 618--629, 2024. doi: 10.1109/ACCESS.2023.3347028

\bibitem{sai2023}
Sai, C. V., Das, D., Elmitwally, N., Elezaj, O., and Islam, M. B. *Explainable AI-Driven Financial Transaction Fraud Detection Using Machine Learning and Deep Neural Networks*. SSRN preprint, 2023. doi: 10.2139/ssrn.4439980


\bibitem{doshi-velez2017}
Doshi-Velez, F., and Kim, B. *Towards A Rigorous Science of Interpretable Machine Learning*. arXiv preprint, 2017. doi: 10.48550/ARXIV.1702.08608

\bibitem{ieeecis2025}
Kaggle. *IEEE-CIS Fraud Detection*. 2025. Accessed: April 23, 2025. Available: \url{https://kaggle.com/ieee-fraud-detection}

\bibitem{talukder2024}
Talukder, Md. A., Hossen, R., Uddin, M. A., Uddin, M. N., and Acharjee, U. K. *Securing Transactions: A Hybrid Dependable Ensemble Machine Learning Model Using IHT-LR and Grid Search*. Cybersecurity, vol. 7, no. 1, p. 32, Nov. 2024. doi: 10.1186/s42400-024-00221-z

\bibitem{yang2019federated}
Yang, Q., Liu, Y., Chen, T., and Tong, Y. *Federated Machine Learning: Concept and Applications*. ACM Trans. Intell. Syst. Technol., vol. 10, no. 2, pp. 1--19, Mar. 2019. doi: 10.1145/3298981

\end{thebibliography}
\end{document}